\documentclass[preprints,article,accept,pdftex,moreauthors]{Definitions/mdpi} 
\firstpage{1} 
\pubvolume{1}
\issuenum{1}
\articlenumber{0}
\pubyear{2026}
\copyrightyear{2026}
\datereceived{ } 
\daterevised{ } 
\dateaccepted{ } 
\datepublished{ } 
\hreflink{https://doi.org/} 

\Title{UMLoc: Uncertainty-Aware Map-Constrained Inertial Localization with Quantified Bounds}

\TitleCitation{UMLoc: Uncertainty-Aware Map-Constrained Inertial Localization with Quantified Bounds}

\Author{Mohammed S. Alharbi \orcidA{} and Shinkyu Park *\orcidB{}}

\AuthorNames{Mohammed S. Alharbi and Shinkyu Park}

\isAPAStyle{%
       \AuthorCitation{Lastname, F., Lastname, F., \& Lastname, F.}
         }{%
        \isChicagoStyle{%
        \AuthorCitation{Lastname, Firstname, Firstname Lastname, and Firstname Lastname.}
        }{
        \AuthorCitation{S. Alharbi, M.; Park, S.}
        }
}

\address{%
\quad \,\, The authors are with the Electrical and Computer Engineering Department, King Abdullah University of Science and Technology (KAUST), Thuwal 23955, Saudi Arabia; Mohammed.alharbi.3@kaust.edu.sa; shinkyu.park@kaust.edu.sa}

\corres{Correspondence: shinkyu.park@kaust.edu.sa}

\abstract{Inertial localization is particularly valuable in GPS-denied environments such as indoors. However, localization using only Inertial Measurement Units (IMUs) suffers from drift caused by motion-process noise and sensor biases. This paper introduces Uncertainty-aware Map-constrained Inertial Localization (UMLoc), an end-to-end framework that jointly models IMU uncertainty and map constraints to achieve drift-resilient positioning. UMLoc integrates two coupled modules: (1) a Long Short-Term Memory (LSTM) quantile regressor, which estimates the specific quantiles needed to define $68\%$, $90\%$ and $95\%$ prediction intervals serving as a measure of localization uncertainty and (2) a Conditioned Generative Adversarial Network (CGAN) with cross-attention that fuses IMU dynamic data with distance-based floor-plan maps to generate geometrically feasible trajectories. The modules are trained jointly, allowing uncertainty estimates to propagate through the CGAN during trajectory generation. UMLoc was evaluated on three datasets, including a newly collected $2$-hour indoor benchmark with time-aligned IMU data, ground-truth poses and floor-plan maps. Results show that the method achieves a mean drift ratio of $5.9\%$ over a $70\,\si{\meter}$ travel distance and an average Absolute Trajectory Error (ATE) of $1.36\,\si{\meter}$, while maintaining calibrated prediction bounds. The code and datasets are available at \href{https://github.com/m9alharbi/umloc.git}{github.com/m9alharbi/umloc.git}.}

\keyword{Inertial Localization, Uncertainty Estimation, Map-Aided Localization, Quantile Regression, Generative Adversarial Networks (GAN), Neural Networks.} 

\usepackage[percent]{overpic}
\usepackage{siunitx}

\begin{document}

\section{Introduction}\label{Section 1}

Indoor localization has gained significant attention in recent years due to its wide range of applications in navigation, healthcare monitoring, emergency response and robotics \cite{Arnold2021}. Existing methods primarily depend on WiFi and Bluetooth, necessitating dense infrastructure, or on LiDAR and camera-based sensors, which require well-lit and feature-rich environments. These methods can achieve high accuracy but are costly, difficult to scale in real-world deployments \cite{ahmetovic2016navcog, onkarpathak2014wi}, power-hungry and may interfere with human activity or raise privacy concerns \cite{zhang2014loam, zhang2015visual, xu2021fast, zhao2020masked}. A promising alternative is the Inertial Measurement Units (IMUs). They are widely available in smartphones, require no line-of-sight, operate reliably across diverse environments and are both energy and computation-efficient. 

Traditional inertial localization methods are typically divided into two categories: Strapdown Inertial Navigation Systems (SINS) and step-based approaches. SINS employ physics-based models to estimate device orientation and integrate linear acceleration to infer position \cite{jimenez2010indoor}. Step-based methods exploit the periodic structure of human gait, relying on hand-crafted features or biomechanical models to approximate pedestrian motion through step detection, heading estimation and step length modeling \cite{jimenez2009comparison, Basso2017}. While conceptually straightforward, these methods are highly susceptible to drift due to the accumulation of integration errors over time. Nonetheless, achieving accurate localization using only IMUs remains fundamentally challenging because of uncertainties in measurement and process, including unmodeled dynamics, sensor biases and the lack of absolute references, e.g., GPS. 

In recent years, deep learning models have shown significant potential in mitigating drift in inertial positioning problems \cite{chen2024deep}. For instance, the Robust IMU Double Integration (RIDI) framework adopts a two-stage pipeline, employing a Support Vector Machine (SVM) to estimate device orientation and a Support Vector Regressor (SVR) to predict velocity vectors in the device frame \cite{yan2018ridi}. Similarly, the Inertial Odometry Network (IONet), a Bidirectional Long Short-Term Memory (LSTM) model, addresses drift by estimating changes in velocity magnitude and direction from accelerometer and gyroscope data, thereby eliminating one level of integration and reducing cumulative error \cite{Chen2018}. Extending this line of work, Robust Neural Inertial Navigation (RoNIN) introduces a robust velocity loss function and investigates multiple network architectures, including LSTM, Temporal Convolutional Network (TCN) and ResNet, to more accurately capture ground-plane motion \cite{herath2020ronin}. Nevertheless, unmodeled uncertainties can limit the model's performance and accumulate errors, leading to trajectory drift.

Tight Learned Inertial Odometry (TLIO) pushes the boundaries of the field by utilizing a ResNet18 model and integrating it with a stochastic-cloning Extended Kalman Filter (EKF). This approach enables the simultaneous estimation of position, orientation, and sensor biases \cite{liu2020tlio}. Similarly, Inertial Deep Orientation-Estimation and Localization (IDOL) combines an LSTM with an EKF, enabling the model to learn velocity changes while using the filter to potentially rectify inaccurate orientation estimates \cite{Sun2021}. Complementing this, Robust Neural Inertial Navigation Aided Visual-Inertial Odometry (RNIN-VIO) tightly couples an EKF-based VIO with a ResNet–LSTM that regresses displacements and their covariance to maintain robustness in weak-texture and motion-blurred scenes \cite{chen2021rnin}. However, these methods modeled uncertainty by imposing a Gaussian distribution rather than providing calibrated prediction intervals. Although prior data-driven methods have achieved promising results, they lack explicit mechanisms to enforce spatial feasibility, often producing trajectories that intersect with obstacles, thus compromising realism and limiting practical deployment.

Despite these advances, existing inertial localization methods still suffer from inevitable drift over time. This limitation motivates exploring auxiliary information, such as map-based constraints, to provide absolute references for correcting drift and improving localization accuracy. Particle filters are widely known for integrating positional estimates with environmental constraints \cite{lui1998sequential}. These filters propagate multiple hypotheses of location based on odometry, retaining or discarding particles based on their consistency with the map and sensor observations. While they are effective in practical applications \cite{xia2018indoor, beauregard2008indoor}, particle filters are highly sensitive to odometry noise, leading to premature removal of valid hypotheses. To address this issue, \cite{melamed2022learnable} introduced learnable map embeddings that adaptively reweight particle distributions, enhancing robustness against noisy motion estimates. However, this method cannot propagate map-consistency violations back into the feature learning process, hence limiting further improvements in localization accuracy.

The lack of an accurate inertial localization model that explicitly incorporates the spatial constraints and uncertainty bounds into position estimation motivates the design of an end-to-end localization framework to address the limitations of existing methods. In this work, we introduce UMLoc, an Uncertainty-aware Map-constrained inertial Localization framework that aims to generate spatially feasible trajectories with quantified bounds. This elevates the task from unconstrained inertial tracking to map-compliant localization with probabilistic guarantees. The proposed architecture consists of two main components: (1) A quantile regression module based on a Long Short-Term Memory (LSTM) network, which estimates prediction intervals on velocity uncertainty. (2) A Conditional Generative Adversarial Network (CGAN) with a cross-attention mechanism, which unifies quantile-derived uncertainty and distance-based map feasibility for trajectory generation. The key innovation lies in its structured conditioning via a cross-attention mechanism, enabling the model to simultaneously generate spatially consistent pedestrian positions within a global reference frame while providing a confidence measure that offers both robustness and interpretability. To achieve our objectives, a curriculum learning is used to optimize the multitasking framework. Starting with an adapted pinball loss for predictive bounds. Then, an adversarial loss and a supervised trajectory loss with a map-compliance penalty are used to train the CGAN for map-compliant trajectory prediction. The proposed UMLoc is extensively evaluated across three datasets, our own dataset and two publicly available datasets. Furthermore, UMLoc is compared against strong baselines, including RoNIN-TCN, RoNIN Bi-LSTM \cite{herath2020ronin} and RNIN \cite{chen2021rnin}, demonstrating superior accuracy, robustness and generalization.  The main contributions of this work are as follows:
\begin{itemize}
    \item We introduce an LSTM-based quantile prediction module that estimates prediction intervals on the velocity, which is integrated to form a positional bound for position estimates.
    
    \item We design CGAN with a cross-attention mechanism to generate trajectories that are consistent with both the learned uncertainty bounds and a 2D environmental map, thereby ensuring map-compliant localization.  

    \item We couple these two modules into a novel end-to-end learning framework for inertial localization that:  
    (1) effectively propagates IMU uncertainty through to the final trajectories and  
    (2) yields reliable, drift-reduced position predictions.  
\end{itemize}
Beyond these technical contributions, we also establish an indoor localization benchmark dataset consisting of synchronized IMU data, ground-truth poses and floor-plan maps.

This paper is organized as follows: Section \ref{Section 2} describes the formulation of the localization problem. Section \ref{Section 3} introduces the proposed UMLoc and its data preparation and training process. Then, the inertial localization results and the ablation study are presented in Section \ref{Section 4}. Finally, concluding remarks on the proposed methods are presented in Section \ref{Section 5}.

\section{Problem Description}\label{Section 2}
We formulate the inertial localization problem as a learning task. It exploits IMU data from a handheld device (e.g., a smartphone) and a corresponding 2D map of the environment to predict the pedestrian's position. Let $X_{1:t}=( x_1,\dots,x_{t} )$ denote IMU measurements from the initial time to the current time $t$, where each vector $x_\tau = (a_{x, \tau},a_{y, \tau},a_{z,\tau},\omega_{x,\tau},\omega_{y,\tau},\omega_{z,\tau})^\top\in\mathbb{R}^6$ at time $\tau \in \{1, \dots, t \}$ contains three-axis linear acceleration $(a_{x,\tau},a_{y,\tau},a_{z,\tau})$ and three-axis angular velocity $(\omega_{x,\tau}, \omega_{y,\tau}, \omega_{z,\tau})$.

The goal is to estimate the corresponding sequence of positions $P_{1:t} = (p_1, \dots, p_{t} )$ within a time window, while accounting for drift and spatial constraints. To facilitate this, we define the velocity sequence as  $V_{1:t} = (v_1, \dots, v_{t} )$, where each velocity vector $v_\tau = (v_{x,\tau}, v_{y,\tau})^\top \in \mathbb{R}^2$ represents motion in the 2D plane. The corresponding position at time $\tau$ is denoted by $p_\tau=(p_{x,\tau},p_{y,\tau})$, lying in a global 2D coordinate frame that is aligned with the environmental map $\mathcal{M}$. We assume that the vertical position $p_{z,\tau}$ remains constant and that the map $\mathcal{M}$ is known a priori. To address the problem, we decompose the localization task into the two interconnected sub-tasks. 

\subsection{Uncertainty Estimation via Quantile Regression}
Let $\mathbf{v}_t$ denote a random variable representing the ground-truth velocity at time $t$. For each time step $t$, the probability that $\mathbf{v}_t$ falls within the conditional lower and upper quantiles, given IMU observation $X_{1:t}$, is defined as: 
\begin{equation} \label{eq:quantile}
\mathbb{P} \left[ q_t^{L} \leqslant \mathbf{v}_t \leqslant q_t^{U} \,|\, v_1, X_{1:t} \right] = 1 - 2\alpha,
\end{equation}
where the inequality in \eqref{eq:quantile} is applied element-wise, and $q_t^{L}, q_t^{U} \in \mathbb R^2$ represent the conditional lower and upper quantiles of $\mathbf{v}_t$ conditioned on the initial velocity $v_1$ and IMU data $X_{1:t}$ with $\alpha \in (0, 0.5)$ denotes the tail probability. These quantiles define the prediction interval at confidence level $1 - 2\alpha$, expressed as $[q_t^{L}, q_t^{U}]$.

To estimate these quantiles, we train a quantile regression network:
 \begin{equation}\label{eq:LSTM}
\left(\hat{q}_t^{L}, \, \hat{q}_t^{U} \right) = \mathcal{F}_{q}(v_1, X_{1:t}; \theta_q),
 \end{equation}
where $\mathcal{F}_{q}$ denotes a quantile regression network parameterized by $\theta_q$, trained using the adapted pinball loss described in Section \ref{Section 32}. 
The predicted quantiles define a prediction interval $[\hat {q}^L_{t}, \hat {q}^U_{t}]$ such that $\hat{\mathbb{P}} [\hat{q}^L_{t} \leqslant \mathbf{v}_t \leqslant \hat{q}^U_{t} ] = 1-2\alpha$. These quantiles capture both model uncertainty and intrinsic data variability, serving as constraints for subsequent trajectory generation. By providing statistically grounded uncertainty margins, the learned bounds improve the reliability and robustness of inertial localization.

\subsection{Map‑Constrained Trajectory Generation}
A generative neural network model is designed to generate realistic velocity trajectories by sampling from the distribution $p_\mathcal{G}$:
\begin{equation}\label{eq:generator_dist}
{\hat{v}}_t \sim p_\mathcal{G} (v_t \mid y_t), \qquad y_t:=\{I_{1:t}, \mathcal{M}, \hat q^{L}_{t}, \hat q^{U}_{t}\},
\end{equation}
where $y_t$ is the condition, $I_{1:t}$ represents IMU features from initial time to current time $t$ and $\mathcal{M}$ encodes the environmental map. The prediction interval $\hat q^{L}_{t}$ and $\hat q^{U}_{t}$ act as quantile-based quantified uncertainty bounds that guide trajectory generation. Subsequently, a multihead cross-attention is employed to fuse the quantile-derived prediction interval and distance-based map feasibility.

\section{Uncertainty-Aware Map-Constrained Inertial Localization}\label{Section 3}
\begin{figure*}[t!]
    \centering
    \includegraphics[width=1.0\linewidth]{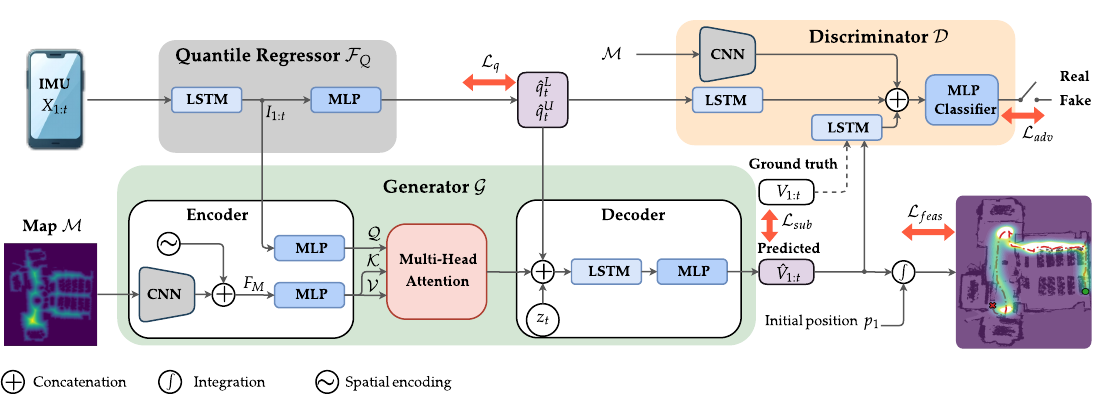}
    \centering
    \caption{Schematic of UMLoc. IMU sequences $X_{1:t}$ feed an LSTM quantile regressor that predicts lower and upper conditional quantiles, while CNN encodes the distance map $\mathcal{M}$. Cross-attention then fuses the IMU with the map features and the decoder generates velocities $\hat{V}_{1:t}$ which are integrated to obtain the positions. The discriminator $\mathcal{D}$ uses the CNN and $2$ LSTMs encoders followed by MLP to distinguish between real and generated sequences.}\label{fig:UMLoc}
\end{figure*}

UMLoc is a two-stage framework that: (1) estimates predictive uncertainty bounds using an LSTM-based quantile regression module and (2) generates spatially feasible velocity trajectories via a CGAN conditioned on an environmental map. Figure~\ref{fig:UMLoc} illustrates the end-to-end pipeline of the proposed UMLoc framework.

We begin by introducing the quantile prediction module, followed by the map-constrained CGAN, which employs a cross-attention mechanism to generate trajectories consistent with both the predicted uncertainty intervals and the environmental map. We then describe the data-collection process used for training and present the progressive training curriculum designed to optimize the model end-to-end.

\subsection{LSTM-based Quantile Regression Module}\label{Section 31}
Localization drift primarily stems from propagated modeling uncertainties, IMU sensor biases and variability in human motion patterns. Instead of relying on deterministic velocity estimates, as in prior work \cite{herath2020ronin, Sun2021, Chen2018}, we employed a quantile regression approach to estimate prediction intervals, providing interpretable measures of uncertainty. 

Formally, given an IMU sequence $X_{1:t}$, the quantile regression module computes the lower and upper quantiles, $\hat{q}^L_{t} = (\hat{q}^L_{x, t}, \hat{q}^L_{y, t})$ and $\hat{q}^U_{t} = (\hat{q}^U_{x,t}, \hat{q}^U_{y,t})$, of the ground-truth $\mathbf{v}_t = (\mathbf{v}_{x,t}, \mathbf{v}_{y,t})\in\mathbb{R}^2$. It is conditioned on the initial velocity $v_1$ and the IMU data $x_1, \dots, x_t$, at each time $t$. To compute these quantiles, the module is trained using an adapted pinball loss. The loss extends the classic formulation \cite{koenker1978regression} by the cumulative sum of the predicted velocity quantiles to estimate positional displacement over the interval $[1,T]$:
\begin{align}\label{eq:apinball}
\mathcal L_{q}&=\frac{1}{4T} \sum_{t=1}^{T} \Bigg[\rho_{\alpha} \bigg(\sum_{k=1}^{t}(\mathbf{v}_{x,k} - \hat{q}^L_{x,k}) \bigg) +\rho_{1-\alpha} \bigg(\sum_{k=1}^{t}(\mathbf{v}_{x,k} - \hat{q}^U_{x,k}) \bigg) \Bigg] \nonumber \\
&\qquad + \frac{1}{4T} \sum_{t=1}^{T} \Bigg[\rho_{\alpha} \bigg(\sum_{k=1}^{t}(\mathbf{v}_{y,k} - \hat{q}^L_{y,k}) \bigg) +\rho_{1-\alpha} \bigg(\sum_{k=1}^{t}(\mathbf{v}_{y,k} - \hat{q}^U_{y,k}) \bigg) \Bigg],
\end{align}
where $T$ is the trajectory length and $\rho_\alpha(u)=\max(\alpha u,(\alpha-1)u)$.

We used a stacked two-layer unidirectional LSTM to extract sequential features $I_{1:t} = \mbox{LSTM}(x_{1:t})$, followed by a Multi-Layer Perceptron (MLP) as a Quantile Regressor $\mathcal{F}_q$ shown in Figure~\ref{fig:UMLoc}. $\mathcal{F}_q$ produces four scalar outputs per time step representing the predicted lower and upper quantiles for the two velocity components, $v_x$ and $v_y$. In this work, we experimented with different values of $\alpha = (0.16, 0.05, 0.025)$, corresponding to $(68\%, 90\%, 95\%)$ prediction intervals, respectively. The resulting interval $(\hat q_t^L, \hat q_t^U)$ serves as an uncertainty-aware constraint, which conditions the trajectory generation module in the second stage of UMLoc.

\subsection{Conditional Generative Adversarial Network (CGAN)}\label{Section 32}

The CGAN module consists of two components: a generator network $\mathcal{G}$ and a discriminator network $\mathcal{D}$, as illustrated in Figure~\ref{fig:UMLoc}. The generator $\mathcal{G}$ learns to produce realistic velocity trajectories by sampling from the learned distribution $p_\mathcal{G}$ as in \eqref{eq:generator_dist}. The discriminator $\mathcal{D}$ learns to classify real and generated velocity trajectories. The overall CGAN objective integrates adversarial loss, supervised estimation loss with map feasibility penalty to jointly improve localization accuracy, spatial feasibility and drift resilience.

The generator $\mathcal{G}$ architecture is shown in Figure~\ref{fig:UMLoc} and consists of a CNN–attention encoder and an LSTM-based decoder. The CNN encoder includes convolutional layers, pooling layers, batch normalization and Rectified Linear Unit (ReLU) activation. The encoder processes the 2D environmental map $\mathcal M\in\mathbb{R}^{H\times W}$ to generate a feature map. Then, it is augmented by the 2D coordinates (spatial encoding) to provide spatial awareness as $F_M\in\mathbb{R}^{H_r\times W_r\times (64 + 2)}$. 

For the multi-head cross attention module, the feature map $F_M$ is projected by the MLP to obtain the keys and values $\mathcal{K}, \mathcal{V} \in \mathbb{R}^{(H_rW_r)\times d}$. The temporal IMU features $I_{1:t}$ are projected to the key attention dimension $d$ through another MLP to form the temporal queries $\mathcal{Q}_t\in\mathbb{R}^{T\times d}$. The multi-head cross-attention mechanism then integrates the temporal features with spatial map features. It enables the model to selectively focus on relevant spatial regions and produce a context vector $c_t$ at the current time:
\begin{equation*}
    c_t = \mathrm{softmax}\left(\frac{(\mathcal{Q}_{t}W_q)(\mathcal{K}W_k)^\top}{\sqrt{d}}\right)(\mathcal{V}W_v),
\end{equation*}
where $W_q$, $W_k$ and $W_v$ are learnable projection matrices.

Overall, this design grounds the IMU sequence in the environment's spatial context while enforcing feasibility constraints that prevent information leakage from invalid areas. To introduce controlled stochasticity and enhance diversity, the context vector $c_t$ is concatenated with latent Gaussian noise $z_t \sim \mathcal N(0,I)$. The decoder then generates the velocity estimate ${\hat{v}}_t$ conditioned on the predicted quantiles $(\hat{q}^L_t, \hat{q}^U_t )$ and context vector ensuring that the generated trajectories remain within the learned quantified uncertainty bounds as follows:
   \begin{equation}
       \hat{v}_t = dec(z_t, c_t, \hat{q}^L_t, \hat{q}^U_t; W_{dec}).
   \end{equation}
 The positions follow discrete integration: $\hat{p}_{t}=\hat{p}_{t-1}+\Delta t\,\hat{v}_t$, where $\Delta t$ is the sampling period which corresponds to $0.0167\, \si{\second}$ in our experiments. 

To differentiate between real velocity trajectory $V_{1:t}$ and generated trajectory $\hat{V}_{1:t}$ the discriminator $\mathcal{D}$ is conditioned on $\{\mathcal{M}, \hat q^{L}_{t}, \hat q^{U}_{t}\}$. Empirically, adding $I_{1:t}$ to $\mathcal{D}$ reduced stability and did not improve the framework's performance. Thus, the discriminator $\mathcal D$ consists of $2$ LSTM encoders to process the trajectories and quantiles along with a CNN to extract a feature vector of the 2D map $\mathcal{M}$. Then, the MLP classifier uses the feature vector and the LSTM encoders' last hidden states to distinguish between real and generated velocity trajectories. It is trained to minimize the following adversarial loss $\mathcal{L_D}$:
\begin{equation}
    \mathcal L_{\mathcal{D}} = -\mathbb{E}_{\hat{v}_t \sim p_\mathcal{G}(v_t | y_t)}[\log(1 - \mathcal D(\hat{v}_t| y_t))] - \mathbb{E}_{v \sim \text{real}}[\log(\mathcal{D}(v_t | y_t))].
\end{equation} 

The generator $\mathcal{G}$ is optimized with a composite weighted objective that combines adversarial loss and a supervised velocity and position loss with a map feasibility penalty: $\mathcal L_\mathcal{G}=\mathcal L_{adv}+\lambda_{feas}\,\mathcal L_{feas} +\lambda_{sup}\,\mathcal L_{sup}$, where
\begin{align*}
   \mathcal{L}_{adv} &= -\mathbb{E}_{\hat{v}_t \sim p_\mathcal{G}(v_t | y_t)} [\log\mathcal{D}(\hat{v}_t|y_t)],\\
\mathcal L_{feas} &=
  \frac{1}{T}\sum_{t=1}^T\Big(\mathrm{max}(R_s - \mathcal{M}(\hat{p}_t), 0)\Big)^2,\\
\mathcal L_{sup} &= \frac{1}{T}\sum_{t=1}^{T} \Big[ \gamma\|p_t - \hat{p}_t \|_{2}^{2} +(1-\gamma) \|v_t-\hat{v}_{t}\|_2^{2}\Big].
\end{align*}
Here, $\mathcal L_{feas}$ penalizes trajectories that are close to or intersect with obstacles in the map. This penalty uses a safety margin, $R_s$, which is applied when the trajectories are within the $R_s$ distance of any obstacles. $R_s$ is chosen empirically to be $0.4\,\si{\meter}$. The supervised loss $\mathcal L_{sup}$ enforces the supervision with $\gamma=0.3$ chosen to balance the velocity and position contribution.  The weights are empirically chosen to balance each learning tasks and stabilizes the training, with $\lambda_{sup}=5$ and $\lambda_{feas}$ linearly ramped over training iterations $i$ to gradually enforce spatial feasibility $\lambda_{feas} = \mbox{min}(0.5, \mbox{max}(0, 0.5\frac{i-10000}{2000}))$.

\subsection{Data Preparation for Training and Evaluation}\label{Section 33}
Each data sample comprises 6-DoF IMU readings collected from an Android smartphone placed in the pedestrian's pants pocket. It consists of 3-axis linear accelerations and 3-axis angular velocities. In addition to ground-truth position data and a 2D map generated using the SLAM API of the Stereolabs ZED~2i camera system, which is fixed to the pedestrian's chest. The IMU measurements are reported in the device body frame $\{b\}$, which changes dynamically as the device moves. In contrast, both the ground-truth positions and the 2D map are represented in a fixed global frame $\{g\}$. It is defined by a right-handed $z$-up coordinate system with the $z$-axis aligned with the gravity. Hence, the Android Game Rotation Vector (GRV) \cite{android2025} is used to estimate the smartphone's orientation $Q_t$ in quaternion form. It is then applied to transform the IMU measurement $x_t$ from the body frame $\{b\}$ to the global frame $\{g\}$: $x_t^{g}= Q_t \, x_t^{b}\, Q_t^{*},$ where $Q_t^{*}$ denotes quaternion conjugate of $Q_t$.

During training, both IMU data and ground-truth velocities are randomly rotated by an angle $\phi \in [0, 2\pi)$ on the horizontal plane \cite{herath2020ronin}. This augmentation compels the model to learn heading-invariant motion patterns such as steps, turns and lateral movements. As a result, the trajectory direction naturally emerges from successive velocity predictions, leading to better generalization at inference. During testing, the GRV is used to estimate the smartphone's orientation to ensure that the $z$-axis of the output frame remains aligned with gravity.

In addition, a 2D map is constructed as a binary occupancy grid $\mathcal{M}_{occ}\in\{0,1\}^{H\times W}$, where static structures such as walls, doors and offices are represented as obstacles. Hence, $\mathcal{M}_{occ}(i,j)$ takes the value $1$ if cell $(i,j)$ is free and $0$ otherwise. To enable differentiability and capture rich spatial context, a Euclidean distance transform $\mathcal{M}_{EDT}$ is applied to the occupancy grid, producing a smoothed representation. In $\mathcal{M}_{EDT}$, each cell stores the shortest Euclidean distance to the nearest obstacle.

The resulting distance values are converted to metric units using the map resolution $r$ (in meters per pixel), yielding the final map representation $\mathcal{M}(i,j)= r\,\mathcal{M}_{EDT}(i,j)$. This continuous map-aware encoding promotes numerical stability, embeds spatial proximity to obstacles and supports gradient-based learning.

\subsection{Training Process}\label{Section 34}
We adopted a curriculum-based learning strategy to meet our objectives of predictive bounds, feasible trajectory generation and reduced drift in inertial localization. The training is conducted in three successive phases to stabilize adversarial learning and propagate uncertainty from quantile bounds into the map-conditioned generator. 

First, the quantile module was pretrained for $150$~epochs by minimizing $\mathcal L_{q}$. Next, CGAN was trained for $50,000$ iterations to generate global velocity sequences, while optimizing $\mathcal L_{\mathcal{G}}$ with the quantile module frozen. Starting with a warm-up supervised learning to minimize $\mathcal L_{sup}$. After that, we annealed the adversarial term $\mathcal L_{adv}$ to shape realistic and multimodal motion once predictions were reasonable. Then, we ramped the feasibility loss $\mathcal L_{feas}$ to enforce map compliance. Finally, both modules are fine-tuned jointly in an end-to-end manner.

We implemented the proposed model using PyTorch $2.1$ and ran it on NVIDIA RTX 2080 Ti GPU ($ 12 GB$). In this work, we tuned the hyperparameters using optuna \cite{optuna_2019} for hyperparameter optimization. For training, the Adam optimizer is used with learning rates of $(0.001, 0.0001, 0.0002)$ for the quantile module, generator and discriminator, respectively. We used the two-time-scale update rule \cite{heusel2017gans}, training the discriminator with a larger learning rate and multiple updates per generator step, which empirically stabilizes CGAN training and improves convergence. A batch size of $16$ and a window of size $120$, corresponding to a $2\,\si{\second}$ are employed to train the proposed model. A scheduler for the learning rate of the quantile module with a reduction factor of $0.75$ when the validation loss does not decrease for $15$ epochs. 

\section{Results}\label{Section 4}
\subsection{Experimental Setup}\label{Section 41}

\subsubsection{Datasets}

We evaluated the proposed UMLoc model on three inertial localization datasets to assess its performance, generalization and robustness. \textbf{RoNIN} \cite{herath2020ronin} and \textbf{RNIN} \cite{chen2021rnin} are two publicly available datasets for inertial localization (see Table~\ref{tab:datasets}). To complement these, we introduce a new dataset, comprising over $2$ hours of pedestrian trajectory data collected indoors. Both IMU and camera data were recorded at a sampling rate of $60\,\si{\Hz}$, with each trajectory lasting no longer than $10$ minutes. We developed a custom Android application to log all sensor streams. This dataset establishes a new benchmark for map-aware inertial localization in indoor environments.

\begin{table}[H]
    \centering
    \caption{Summary of datasets statistics}
    \begin{tabular}{|l|ll|ll|l|}
        \hline 
         \multirow[l]{2}{*}{Dataset} & \multicolumn{2}{c|}{Travel time $(\si{\second})$} & \multicolumn{2}{c|}{Travel distance $(\si{\meter})$} & Frequency \\
         & Total & Mean $\pm$ std & Total & mean $\pm$ std &  $(\si{\Hz})$ \\
         \hline
         RoNIN & 82,332 & 552 $\pm$ 164 & 62,030 & 416 $\pm$ 154 & 200 \\
         RNIN & 25,200 & 246 $\pm$ 318 & 27,820 & 92 $\pm$ 160 & 100 \\
         UMLoc & 7,488 & 125 $\pm$ 65 & 4,050 & 67 $\pm$ 34 & 60\\
         \hline
    \end{tabular}
    \label{tab:datasets}
\end{table}

\subsubsection{Baselines and metrics definitions}\label{Section 52}
To evaluate the proposed inertial localization model UMLoc, we compared its performance against three state-of-the-art baselines:
\begin{itemize}
    \item \textbf{RoNIN LSTM}\cite{herath2020ronin}: A recurrent model employs LSTM layers with bilinear layers to regress velocity directly from IMU data.
    
    \item \textbf{RoNIN TCN}\cite{herath2020ronin}: A convolution-based alternative that utilizes a Temporal Convolutional Network (TCN) for velocity prediction from IMU data.
    
    \item \textbf{RNIN-VIO}\cite{chen2021rnin}: An uncertainty-aware deep learning model that explicitly predicts displacement along with its associated covariance from IMU data.
\end{itemize}
By benchmarking against diverse architectures and datasets, we aim to comprehensively demonstrate the effectiveness and robustness of UMLoc across different environments, people, devices and multiple levels of sensor noise captured in the datasets.

For systematic evaluation, we employ multiple quantitative metrics for each position trajectory of length $T$:
\begin{itemize}
    \item \textbf{Absolute Trajectory Error (ATE):} 
    The Root Mean Square Error (RMSE) between estimated and ground-truth positions is given as $\text{ATE} =\sqrt{ \frac{1}{T} \sum_{t=1}^{T} || \mathbf{\hat{p}}_t - \mathbf{p}_t ||_2^2}$. It reflects global trajectory consistency and accumulates over time due to drift.
    
    \item \textbf{Relative Trajectory Error (RTE):} 
    The RMSE of position differences over a fixed time interval ($\Delta t = 1$ minute) is calculated as $$\text{RTE} = \sqrt{\frac{1}{T-\Delta t}\sum_{t=1}^{T-\Delta t}||(\mathbf{p}_{t+\Delta t} - \mathbf{p}_t)-(\mathbf{\hat{p}}_{t+\Delta t} - \mathbf{\hat{p}}_t)||_2^2}.$$ It captures local trajectory consistency. 
    
    \item \textbf{Final Drift Error (FDE):} The final displacement error is normalized by the total trajectory distance $L$, $FDE = \lVert \mathbf{\hat{p}}_T-\mathbf{p}_T \rVert_2^2/L$. It quantifies long-term drift relative to the path distance.
\end{itemize}

\subsubsection{Smartphone orientation handling}
RoNIN performs pre-alignment of the tracking and IMU devices at the start and end of each recording session \cite{herath2020ronin}. Specifically, it employs two smartphones, one dedicated to collect ground-truth trajectories and the other to capture the pedestrian’s actual motion data. In addition, orientation supervision is applied during training using ground-truth device orientations. However, at test time the model relies on device orientation, making orientation errors a dominant source of drift. RNIN, by contrast, employs controlled setups (e.g., VICON motion capture systems or rigid camera–IMU assemblies) and leverages ground-truth orientations, thereby bypassing orientation noise inherent in free-hand use. However, in real-world scenarios, ground-truth orientations are unavailable. To address this, UMLoc is trained and evaluated exclusively using GRV, which reflects realistic deployment conditions. This design makes our experiments particularly challenging, as no ground-truth orientations are assumed.

\subsection{Experiment Results}\label{Section 42}
For each dataset (RoNIN, RNIN, ours), we adopted a zero-shot protocol by holding out entire sequences from unseen users, devices and buildings for testing. The Zero-shot testing demonstrates performance and generalization to unseen data. Also, in all experiments, we used the $95\%$ prediction interval for the UMLoc model as it provided more robust and accurate localization than the $68\%$ and $90\%$ intervals unless stated otherwise. Table~\ref{tab:acc_performance} reports the evaluation metrics for all the datasets on UMLoc against baseline models to summarize their performance. Across these splits, UMLoc transfers without fine-tuning, maintaining competitive accuracy on new people and phones (a capability particularly demonstrated by the results on the RoNIN and RNIN datasets). Furthermore, cross-building testing in our dataset demonstrates that UMLoc maintains competitive accuracy in diverse layouts.

\begin{table}[t!]
\caption{Performance evaluation for UMLoc against three baseline models RNIN-VIO and RoNIN LSTM (RoNIN-L)/TCN (RoNIN-T) on RoNIN, RNIN and our datasets. Improvement is calculated based on the best baseline model.}
\label{tab:acc_performance}
\centering
RoNIN Dataset \\ \vspace{.5em}
\begin{tabular}{lccc}
\toprule
 Model & FDE ($\%$) & ATE ($\si{\meter}$) & RTE ($\si{\meter}$) \\
\midrule
RNIN & 4.0 & 7.05 & 6.48  \\
RoNIN-L & 4.6 & 8.73  & 4.87  \\
RoNIN-T & 4.3 & 7.27 & 4.27  \\
UMLoc (no map) & \textbf{3.5}  & \textbf{6.58} & \textbf{3.93}   \\
\midrule
Improvement & 12.50\% & 6.67\% & 7.96\% \\
\bottomrule
\end{tabular}
\\ \vspace{1em}
RNIN Dataset \\ \vspace{.5em}
\begin{tabular}{lccc}
\toprule
Model & FDE ($\%$) & ATE ($\si{\meter}$) & RTE ($\si{\meter}$) \\
\midrule
RNIN & \textbf{2.2} & 1.59 & 2.57  \\
RoNIN-L & 8.0 & 2.66 & 3.93  \\
RoNIN-T & 5.2 & 2.72 & 3.90 \\
UMLoc (no map) & 3.4 & \textbf{1.47} & \textbf{2.03}  \\
\midrule
Improvement & - & 7.55\% & 21.01\% \\
\bottomrule
\end{tabular}
\\ \vspace{1em}
Our Dataset \\ \vspace{.5em}
\begin{tabular}{lccc}
\toprule
Model & FDE ($\%$) & ATE ($\si{\meter}$) & RTE ($\si{\meter}$) \\
\midrule
RNIN & 17.2 & 4.65 & 4.61 \\
RoNIN-L & 23.2 & 4.14 & 4.94  \\
RoNIN-T & 19.2 & 3.66 & 4.62   \\
UMLoc (no map) & 12.1 & 2.33 & 3.23  \\
\midrule
Improvement & 29.65\% & 36.34\% & 29.93\% \\
\midrule
UMLoc & \textbf{5.9} & \textbf{1.36} & \textbf{1.91} \\
\midrule
Improvement & 65.70\% & 62.84\% & 58.57\% \\
\bottomrule
\end{tabular}
\end{table}
\subsubsection{Evaluation on RoNIN and RNIN datasets}

RoNIN and RNIN are two widely used indoor inertial benchmark datasets containing diverse pedestrian trajectories \cite{herath2020ronin, chen2021rnin}. These datasets are collected from different devices and by different people. The RoNIN dataset comprises over 40 hours of recordings and uses a $85/15$ train–test split, whereas RNIN includes 7 hours of data with a $90/10$ split, following the original papers' setup. Since neither dataset provides map information, we supplied UMLoc's CNN branch with a \emph{uniform feasibility map} to maintain identical model capacity and ensure a fair comparison that quantifies IMU-only performance.

Table~\ref{tab:acc_performance} shows that UMLoc without map conditioning achieves an ATE of $6.58\,\si{\meter}$ and FDE of $3.5\%$ over the traveled distance on RoNIN datasets. For RNIN it attains ATE of $1.47\,\si{\meter}$ and RTE around $2\,\si{\meter}$ over a period of $1\,\si{\minute}$. These results show marginal improvements across all metrics relative to the strongest baselines (see the ``Improvement'' rows).

\subsubsection{Evaluation on our dataset}

\begin{figure}[H]
\includegraphics[width=\linewidth]{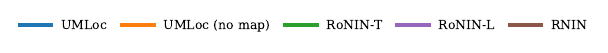}
   \centering
      \begin{overpic}[width=\linewidth]{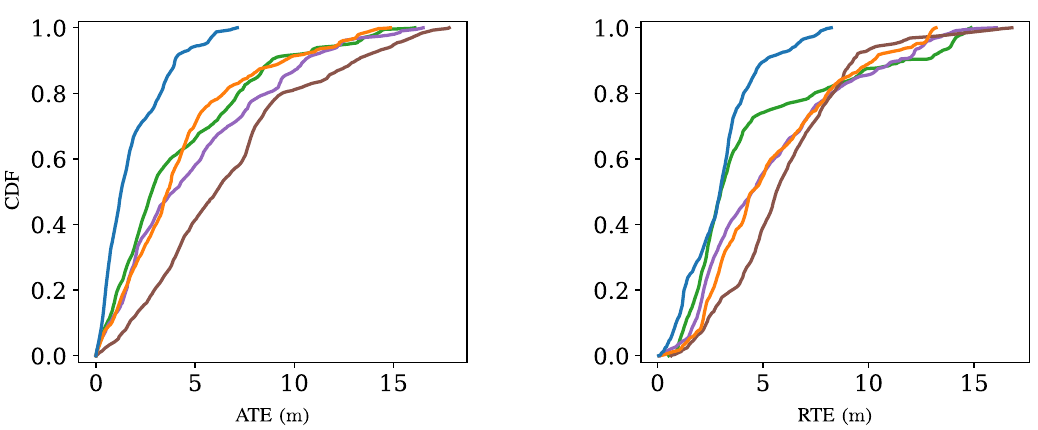}
            \end{overpic} 
            \vspace{0.01em}
       \caption{CDFs of ATE (left) and RTE (right) on our datasets' unseen testing split. UMLoc achieves $80\%$ cumulative probability below $2.5\,\si{\meter}$ error, whereas other models have it at $7.5\,\si{\meter}$.}
       \label{fig:cdf}
       \end{figure}

\begin{figure}[H]
\includegraphics[width=\linewidth, height=0.08\linewidth]{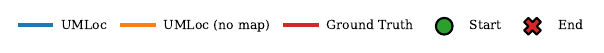}
   \centering
      \begin{overpic}[width=1\linewidth, height=0.67\linewidth]{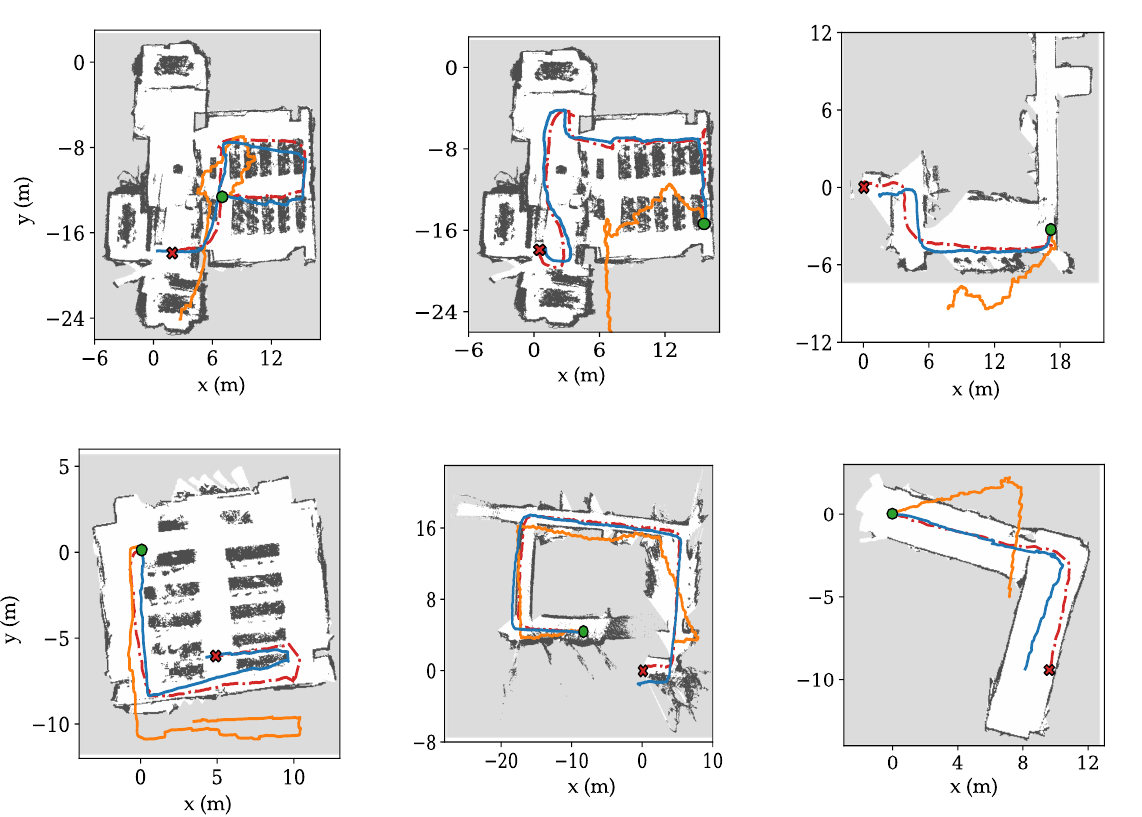}
            \end{overpic} 
       \caption{Selected 2D trajectory visualizations. We selected 6 trajectories from our dataset that include the map.}
       \label{fig:trajectory_viz}
\end{figure}

We split the trajectories into an $85/15$ split for training and testing. All baseline models were retrained using the identical splits to ensure fair comparison. On our dataset, UMLoc with map conditioning outperforms all baselines, as well as its IMU-only variant, achieving over $50\%$ improvement over the strongest baseline. UMLoc achieved an average drift of $5.9\%$ over an average traveled distance of $70\,\si{\meter}$ corresponding to a total combined travel distance over $400\,\si{\meter}$ for testing trajectories (see Table~\ref{tab:acc_performance}). Also, the cumulative distribution functions (CDFs) of ATE and RTE show that UMLoc is the first to reach its maximum error as illustrated in Figure~\ref{fig:cdf}. Figure~\ref{fig:trajectory_viz} shows the visualization of selected unseen trajectories on their maps to compare the UMLoc with its IMU-only variant. It can be shown that UMLoc accurately predicts map-compliant trajectories and effectively avoids obstacle areas.

\begin{figure}[H]
    \centering
     \begin{overpic}[width=1\linewidth, height=0.67\linewidth]{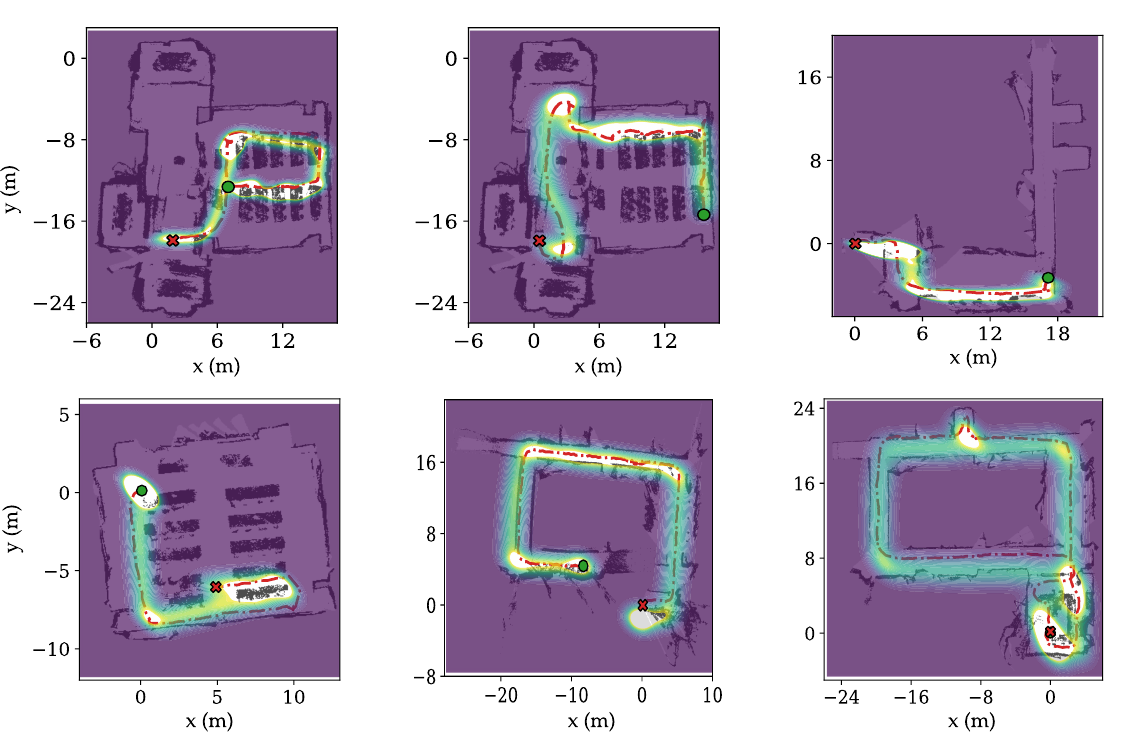}
            \end{overpic}
    \caption{Illustrations of $20$ trajectory samples from CGAN for $6$ testing trajectories from our dataset. High intensity represents a higher probability of pedestrian location. }
    \label{fig:Uncertainity}

\end{figure}

To test the model's ability to generate multiple trajectories that follow the map constraints and to assess its uncertainty estimation, we draw $20$ trajectory samples from the CGAN. In Figure~\ref{fig:Uncertainity}, the density is demonstrated over the 2D map with the ground truth trajectory. The resulting high-density ridge closely follows the ground-truth path while remaining within free space, indicating slight bias and good adherence to map constraints.

\subsubsection{Ablation Study Map Conditioning versus No Map Conditioning}
\begin{figure}[H]
    \centering
     \begin{overpic}[width=.6\linewidth]{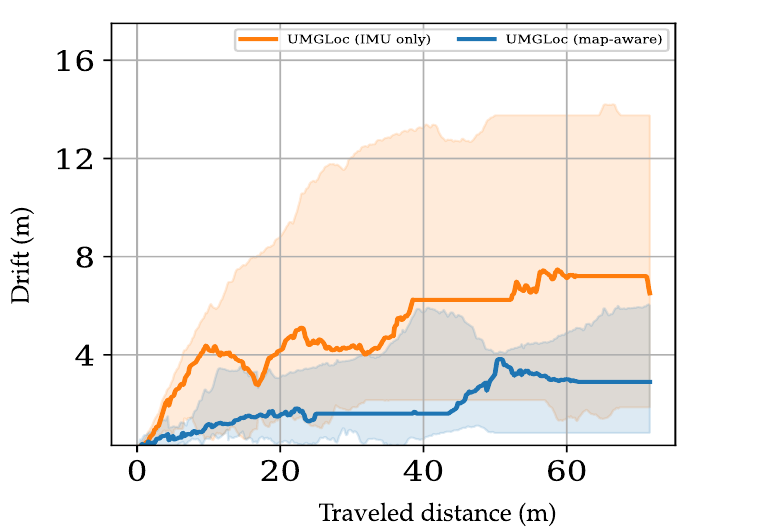}

            \end{overpic}
    \caption{Drift error versus traveled distance for the map-aware model (UMLoc) and its IMU-only variant. Curves show the median drift across unseen test sequences. We computed the error over the average travel distance of $70\,\si{\meter}$ after aggregating all the trajectories. The shaded area denotes the drift distribution over the testing trajectories. Incorporating map information (blue) constrains maximum drift growth to $5\,\si{\meter}$ throughout the testing trajectories versus $13\,\si{\meter}$ in the IMU-only model (orange), demonstrating the map’s effectiveness in long-distance indoor localization.}
    \label{fig:map-no-map}

\end{figure}

We conducted an ablation study to explicitly assess the benefit of incorporating map information into the localization pipeline. Figure~\ref{fig:map-no-map} presents the drift-error analysis that is defined as the position estimation error (i.e., Euclidean distance between the ground truth and predicted positions). The ablation results reveal a substantial difference in drift, validating the synergy between uncertainty bounds and spatial feasibility constraints. Notably, UMLoc with a map consistently exhibits lower cumulative drift than its IMU-only counterpart. The map-aware approach effectively suppresses drift growth, maintaining drift errors to around $5\,\si{\meter}$ compared to $13\,\si{\meter}$ over an average traveled distance of $70\,\si{\meter}$ throughout the entire trajectories. These findings highlight the critical role of map constraints in stabilizing localization performance, substantially reducing positional uncertainty and enhancing reliability over extended deployments.

\subsubsection{Robustness Test}
To evaluate the resilience of the proposed UMLoc pipeline under realistic operating conditions, we introduced two common sensor degradations \emph{only at the testing time}:

\textit{(i)}~\textbf{Additive Gaussian noise.} Each accelerometer and gyroscope readings were corrupted with zero-mean and standard deviations \(\sigma \in \{0,\,0.1\sigma_{IMU},\,0.5\sigma_{IMU},\,1.0\sigma_{IMU},\,5.0\sigma_{IMU}\}\) where $\sigma_{IMU}$ is the captured IMU standard deviation. This perturbation mimics thermal noise in low-cost MEMS units, human-induced vibrations, electromagnetic interference and the gradual increase in noise power caused by sensor aging.

\textit{(ii)} \textbf{Random sample dropout.} $10\%$ of IMU frames were randomly replaced with $0$. This scenario reflects real-world issues such as packet loss on wireless links and power-saving modes.

For the above two settings, we computed the Prediction-Interval Coverage Probability (PICP) and Average Interval Width (AIW) to evaluate the quantile calibration: 
\begin{align} 
\text{PICP}\; &= \frac{1}{T}\sum_{t=1}^{T} \mathbf{1}\!\left[\hat q^L_t \le v_t \le \hat q^U_t \right], \label{eq:picp} \\
\text{AIW}\; &= \frac{1}{T}\sum_{t=1}^{T} \left\lVert \hat q^U_t - \hat q^L_t \right\rVert_{2},
\end{align}
where the inequality in \eqref{eq:picp} is applied element-wise.
Since RoNIN does not natively provide uncertainty measures for a fair comparison, we used RNIN as the baseline for the robustness test. We derived the component-wise prediction intervals of $68 \%$, $90 \%$ and $95 \%$ (corresponding to $1-2\alpha$)  by using multiples of the predicted standard deviation $\hat{\sigma}_t$, specifically $(1 \hat{\sigma}_t$, $1.64 \hat{\sigma}_t$ and $2 \hat{\sigma}_{t})$ (details in \ref{Appendix A}). For example, the $95 \%$ interval bounds were calculated as:
\begin{equation}
\hat q^L_{t}=\hat{v}_{t}-\,2\hat{\sigma}_{t},\quad
\hat q^U_{t}=\hat{v}_{t}+\,2\hat{\sigma}_{t}.
\end{equation}

\begin{figure}[H]
   \centering
   \includegraphics[width=\linewidth, height=0.07\linewidth]{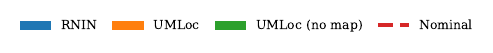}\\
   \includegraphics[width=\linewidth, height=1.1\linewidth]{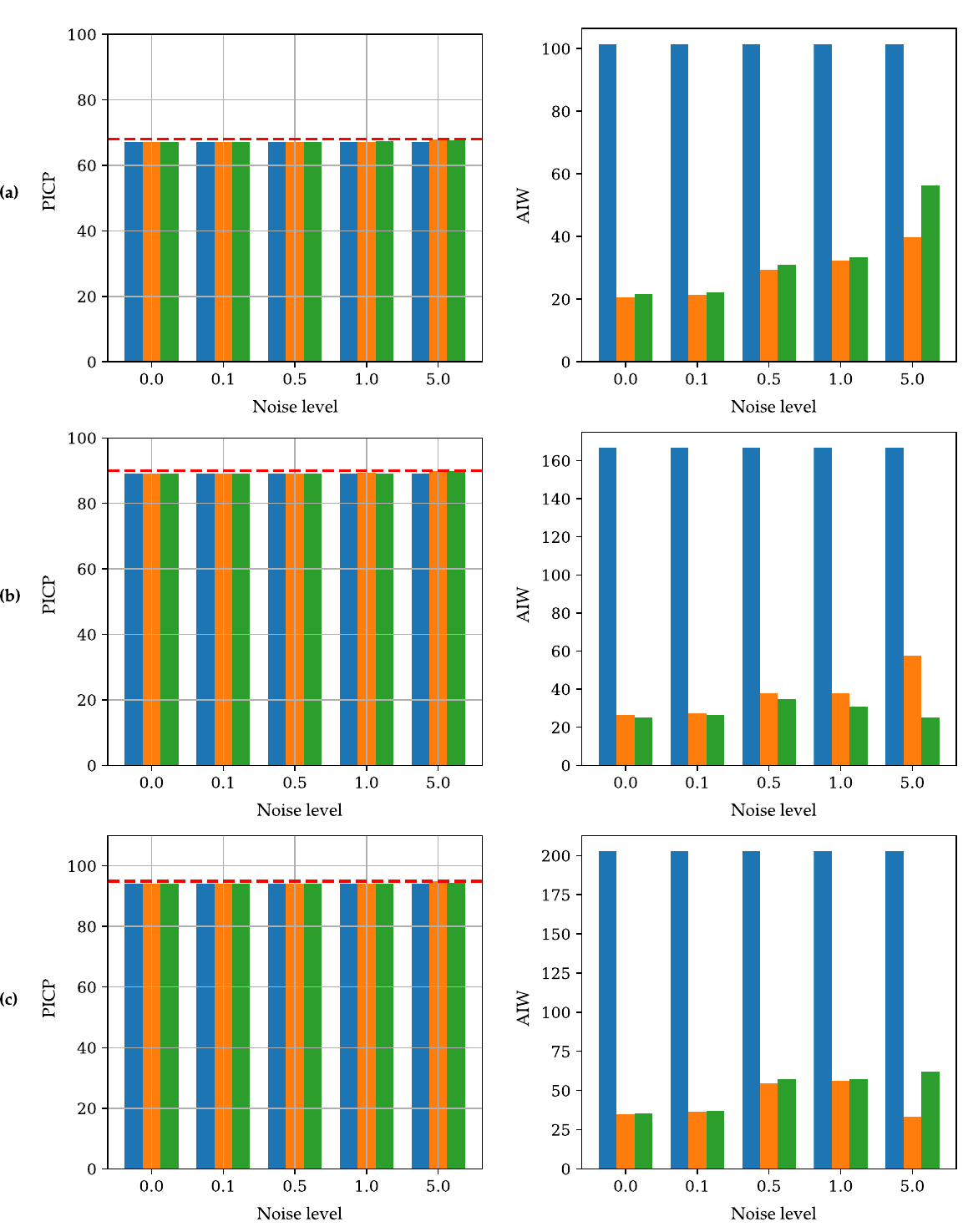}
       \caption{Robustness test of three models under increasing IMU noise and $10\%$ random dropout. On the left is the PICP and AIW is shown on the right, such that \textbf{(a)} is the $68\%$ prediction interval; \textbf{(b)} is for $90\%$ prediction interval; \textbf{(c)} represents the $95\%$ prediction interval. All three models maintain PICP near the nominal level across all noise levels. However, our proposed model, UMLoc, with and without map variants, adaptively widens their prediction interval as the noise level increases. In contrast, RNIN gives a high constant AIW that reflects maintaining PICP close to the nominal.to reflect good coverage. Hence, the plots highlight the superior calibration of the proposed UMLoc model.}\label{fig:robust_test}
       \end{figure}

The evaluation of the prediction interval is shown in Figure~\ref{fig:robust_test}, which reports PICP and AIW for the $3$ intervals $\{68\%, 90\%, 95\%\}$. The figure highlights how well the intervals are maintained against different operating conditions. UMLoc consistently maintains its coverage probability close to the nominal target and exhibits the most adaptive increase in interval width even at high noise levels. While, RNIN maintains coverage, but the intervals are unrealistically wide, reflecting good coverage performance. Overall, UMLoc provides the best balance, effectively maintaining high coverage and a reasonable interval width under noisy conditions, which clearly outperforms RNIN in terms of robustness and calibration. Demonstrating that our method’s explicit uncertainty quantification is robust against the inevitable noise, interference and data gaps encountered in practical deployments.

\subsubsection{Generalization}
The generalizability of UMLoc is tested across two parts. First, the zero-shot testing previously shown in Table~\ref{tab:acc_performance}. Especially, the cross-building testing, where UMLoc maintains competitive accuracy across diverse layouts in our dataset, whereas other models fail in similar cases during cross-building evaluation \cite{Sun2021}. Second, to test our model's generalizability, we trained an independent model for each building in our dataset, while holding out unseen test trajectories from that building. Then, we compared the ATE from a single globally trained UMLoc model with the average ATE from separate, independent models, each trained and tested exclusively on a single building's dataset. The performance comparison is presented in Table~\ref{tab:Gen_test}.

We attribute the strong generalizability of the single UMLoc model to three main factors: (i) the IMU encoder learns kinematic patterns that are agnostic primarily to the users and environments; (ii) device differences are mitigated through noise augmentation and the use of the quantile objective, which encourages calibrated and less over-confident predictions under data shifts; and (iii) the map-aware feasibility prior effectively constrains drift even in unfamiliar spaces, provided the floor plan is well-aligned with the global frame.

\begin{table}
\caption{Generalizability evaluation for UMLoc model using independent models for each building against one model for the combined buildings.}
\label{tab:Gen_test}
\centering
\begin{tabular}{lc}
\toprule
 Model & ATE ($\si{\meter}$) \\
\midrule
Building 1 & 1.12\\
Building 2 & 1.05 \\
Building 3 & 0.89 \\
Building 4 & 1.73 \\
Building 5 & 0.95 \\
Independent models average & 1.15 \\
\midrule
All building model & 1.36 \\
\bottomrule
\end{tabular}
\end{table}
\section{Conclusion}\label{Section 5}
In this paper, we propose UMLoc, a novel framework that integrates an LSTM-based quantile-prediction module with a map-aware CGAN to reduce drift and provide explicit prediction intervals. Through comprehensive experiments across our dataset and benchmark datasets, UMLoc consistently outperforms state-of-the-art approaches. The results demonstrate substantial improvements in accuracy, robustness and generalization capabilities. This work opens new avenues for researchers to explore more sophisticated integration techniques, extend the concept to various sensor modalities and address real-world localization challenges.

Future research directions include extending UMLoc to three-dimensional localization and addressing deployment challenges in real-world indoor navigation (i.e., shopping malls, airports, office buildings, hospitals and industrial facilities) and robotic applications.

\vspace{6pt} 





\authorcontributions{M.S.A. conceptualized and implemented the algorithm, carried out the experiments and the data collection, drafted the initial manuscript. S.P. supervised the research direction, provided resources, revised and approved the final manuscript. All authors have read and agreed to the
published version of the manuscript.}

\funding{The work was supported by funding from King Abdullah University of Science and Technology (KAUST).}

\dataavailability{The dataset and codes used in this work are publicly available on: \href{https://github.com/m9alharbi/umloc.git}{github.com/m9alharbi/umloc.git}.}

\acknowledgments{The authors would like to acknowledge the assistance from the KAUST Facilities and KAUST Supercomputing Laboratory (KSL) for making the training possible.}

\conflictsofinterest{The authors declare no conflicts of interest. The funders had no role in the design of the study; in the collection, analyses, or interpretation of data; in the writing of the manuscript; or in the decision to publish the results.}

\appendixtitles{yes} 
\appendixstart
\appendix
\section[\appendixname~\thesection]{Post quantile calibration for RNIN}\label{Appendix A}
Besides the point estimate \(\hat{v}_{t}\in\mathbb{R}^{2}\),
RNIN outputs a vector
\(\mathbf{c}_{t}\in\mathbb{R}^{D}\) that parametrizes the diagonal
predictive covariance.
Following \cite{chen2021rnin}, each entry is interpreted as the logarithm of the
per–dimension \emph{standard deviation},
\begin{equation}
\hat \sigma_{t}\;=\; \begin{pmatrix} \exp (c_{t,1}) \\ \exp (c_{t,2} ) \end{pmatrix}.
\end{equation}


\reftitle{References}
\isAPAandChicago{}{%
\bibliography{References}}
\PublishersNote{}
\end{document}